\documentclass[runningheads]{llncs}
\usepackage[T1]{fontenc}

\usepackage{amsmath,amssymb,amsfonts}
\usepackage{algorithmic}
\usepackage{graphicx}
\usepackage{textcomp}
\usepackage[table]{xcolor}
\usepackage{tabularx}
\usepackage{xcolor}
\usepackage{array}
\usepackage{multirow}
\usepackage{rotating}
\usepackage{hyperref}
\usepackage{placeins}
\usepackage{soul}

\usepackage{orcidlink}

\begin{document}
\title{Can Vision Transformers with ResNet's Global Features Fairly Authenticate Demographic Faces?}
%
%
\author{Abu Sufian\inst{1,2}\orcidlink{0000-0003-2035-2938} \and
Marco Leo\inst{1}\orcidlink{0000-0001-5636-6130} \and
Cosimo Distante\inst{1}\orcidlink{0000-0002-1073-2390} \and Anirudha Ghosh \inst{3}\orcidlink{0009-0004-1465-0386} \and
Debaditya Barman\inst{3}\orcidlink{0000-0002-7562-119X}}
\authorrunning{A. Sufian et al.}
\institute{National Research Council of Italy - Institute of Applied Sciences and Intelligent Systems (CNR-ISASI), 73100 Lecce, Italy.\\
\email{\{abu.sufian, marco.leo, cosimo.distante\}@isasi.cnr.it} \and University of Gour Banga, English Bazar, 732103, India. \and  Visva-Bharati, Santiniketan, 731235, West Bengal, India. \\
\email{03333342106@visva-bharati.ac.in, debadityabarman@gmail.com}}
\maketitle              
\begin{abstract}
Biometric face authentication is crucial in computer vision, but ensuring fairness and generalization across demographic groups remains a big challenge. Therefore, we investigated whether Vision Transformer (ViT) and ResNet, leveraging pre-trained global features, can fairly authenticate different demographic faces while relying minimally on local features. In this investigation, we used three pre-trained state-of-the-art (SOTA) ViT foundation models from Facebook, Google, and Microsoft for global features as well as ResNet-18. We concatenated the features from ViT and ResNet, passed them through two fully connected layers, and trained on customized face image datasets to capture the local features. Then, we designed a novel few-shot prototype network with backbone features embedding. We also developed new demographic face image support and query datasets for this empirical study.  The network's testing was conducted on this dataset in one-shot, three-shot, and five-shot scenarios to assess how performance improves as the size of the support set increases. We observed results across datasets with varying races/ethnicities, genders, and age groups. The Microsoft Swin Transformer backbone performed better among the three SOTA ViT for this task. The code and data are available at: https://github.com/Sufianlab/FairVitBio.

\end{abstract}
\section{Introduction}
\label{intro}
Over the past decade, deep learning (DL) \cite{lecun2015deep,prince2023understanding} have demonstrated remarkable performance across various domains, including computer vision (CV) \cite{voulodimos2018deep}, natural language processing (NLP) \cite{deng2018deep,khan2023exploring}, biometrics \cite{minaee2023biometrics,das2023sclera} and more. The advent of Transformer Architecture \cite{NIPS2017_3f5ee243}, Vision Transformer (ViT) \cite{ViT,liu2023survey}, Large Language Models (LLMs) \cite{radford2019language,chang2024survey,minaee2024large}, and Large Vision Language Models (LVLMs) \cite{bordes2024introduction,zhang2024vision} have further strengthened the backbone choices for many applications including biometric face authentication. However, this rapid development has come at a significant cost: an increased demand for computational resources, such as reliance on vast training data, often leads to fairness issues and biases \cite{drozdowski2021watchlist,nazer2023bias}. These limitations also hinder the practicality of an AI model in resource-constrained environments, especially for face authentication applications that require efficient and scalable solutions, where minimizing data-driven biases is crucial   \cite{drozdowski2021watchlist,khan2023exploring,chang2024survey}. In recent times, transfer learning and fine-tuning of LLM and LVLM have emerged as promising strategies to alleviate these issues by reducing the need for large datasets and enabling models to generalize more effectively across downstream tasks \cite{thirunavukarasu2023large,wang2023pre}. However, even these methods struggle to build a generalized model to classify high-similarity images efficiently, especially in biometric face authentication tasks, where fairness and biases are high. Moreover, database demographic biases result in inequitable performance across different demographic groups of people \cite{valdivia2023there,sanon2024study}.

To contribute to addressing these challenges, we empirically studied the hypothesis: Can Vision Transformers with ResNet's Global Features Fairly Authenticate Demographic Faces? We explored the use of ViT \cite{ViT,liu2023survey} and ResNet's \cite{he2016deep} global features, with minimal reliance on local features, for backbone model development. We added two fully connected (FC) layers to the pre-trained ViTs and ResNet, fine-tuning them using our customized training dataset to extract local features. We designed a novel few-shot prototype meta-learning network with this backbone for this study. We experimented with it on a customized face image dataset from three reliable sources. 
We then tested the final trained prototype meta-learning network with our created demographic face datasets separately with the three ViT models: Microsoft Swin Transformer (ST) \cite{swin}, Facebook DeiT \cite{touvron2021training,yao2024data}, and Google VT \cite{wu2020visual}. We systematically compare the results with each SOTA ViT backbone through the prototype meta-learning networks. The highlights of this paper include:
\begin{itemize}
	\item We conducted a novel empirical study on three SOTA ViT backbones in the demographic face authentication task.  
	\item We designed an innovative backbone architecture using pre-trained ViT and ResNet to use global features and two FC layers for local features. 
	\item We introduced a novel few-shot prototype meta-learning network to evaluate the performances. 
	\item We created a new demographic face authentication dataset using an API.  
	\item We analyzed the performances of each of the three backbone models on our demographic face datasets and found that the Microsoft ST ViT backbone performed comparatively better. 
        \item We did an ablation study to understand the performance of ViT with and without ResNet.   
\end{itemize}
The remainder of this paper is structured as follows: Section \ref{LR} provides a briefing on the literature review. Section \ref{Method} presents the methodology. Experimental details are outlined in Section \ref{Exp}, followed by the presentation of results and discussion in Section \ref{RD}. The ablation study can be found in Section \ref{Ablation}, whereas the conclusion and implications of future works are in Section \ref{Conclusion}. 

\section{Literature Review}
\label{LR}
\textbf{Visual Transformer (ViT)}:
ViT \cite{ViT} is a transformer-based approach adapted in CV, originally designed for NLP \cite{NIPS2017_3f5ee243}. Unlike Convolutional Neural Networks (CNNs) \cite{ghosh2020fundamental}, which focus on local patterns at each convolution, ViTs treat images as sequences of fixed-size patches. These patches are embedded into high-dimensional vectors and processed by a transformer encoder through self and cross-attentions, allowing ViTs to capture long-range dependencies and global context more effectively than CNNs. These strengths enable ViTs to model relationships across the entire image as global features, making them powerful models. \\ 
Though ViTs generally require large datasets for optimal performance, they have shown superior results compared to CNNs across various benchmarks \cite{wu2020visual,liu2023survey}. The ST model \cite{swin} introduces hierarchical feature maps and shifted windows for improved efficiency. In contrast, Data-efficient Image Transformers (DeiT) model \cite{touvron2021training,yao2024data} enhances data efficiency through a teacher-student training strategy. Pyramid ViT (PVT) model \cite{wang2021pyramid} uses a pyramid structure for better handling of high-resolution images, and Convolutional ViT (CvT) model \cite{wu2021cvt} integrates CNN layers into the transformer. These advancements extend ViTs' adaptability to a broader range of CV tasks. \\ \\
\textbf{Few-Shot Meta-Learning}:
Few-shot meta-learning or simply FSL is an ML technique designed to handle tasks with limited labeled training data, allowing models to generalize and make predictions from a small set of examples \cite{sung2018learning,wang2020generalizing}. By leveraging transfer learning (TL) and meta-learning, FSL enables models to adapt to new tasks with minimal training. This makes it useful in areas like biometric face authentication, where data scarcity is common \cite{song2023comprehensive}.\\
FSL has proven effective in face authentication, offering solutions for data scarcity while maintaining high accuracy \cite{browatzki20203fabrec,holkar2022few}. It addresses the need for generalization and personalization in face authentication the approaches like prototype networks \cite{snell2017prototypical}, Siamese networks \cite{koch2015siamese}, and Model-Agnostic Meta-Learning (MAML) \cite{finn2017model} are highly effective. In this work, we design a few-shot prototype meta-learning network \cite{snell2017prototypical} using ViT and ResNet backbone to handle demographic face authenticate datasets.\\ \\
\textbf{Face Authentication}:
Face authentication uses unique facial features to verify identity and is widely used in applications like biometric attendance and security systems \cite{imaoka2021future,sufian2023fewfacenet,davis2024super}. Despite the task looks fairness and biases and fairness remain a key challenge \cite{alasadi2019toward,valdivia2023there,fang2024fairness}. Many systems suffer from biases due to unrepresentative training data, leading to higher error rates for specific demographic groups based on race, gender, and age \cite{drozdowski2021watchlist,de2021fairness,chandaliya2024towards}. These issues raise ethical concerns and affect fair accuracy, highlighting the need for fair solutions. \\
ViTs have shown promise in improving fairness and accuracy by capturing relevant global features better than CNNs \cite{luo2022memory,kim2023s,abdrakhmanova2024one}. ViTs can analyze both global and local features, making them well-suited for handling diverse datasets and reducing bias. Combining ViTs with bias mitigation techniques and fairness metrics during model development can enhance performance across demographics. Researchers are also developing diverse datasets and rigorously testing models for fairness \cite{yang2024demographic,andrews2024ethical,shahreza2024sdfr,leschanowsky2024data}. While ViTs with FSL show potential in biometrics, continuous research is needed to address demographic biases. 
\section{Methodology}
\label{Method}
\subsection{The ViT Backbone with ResNet}
\begin{figure}
	\begin{center}
		\includegraphics[width=0.99\linewidth]{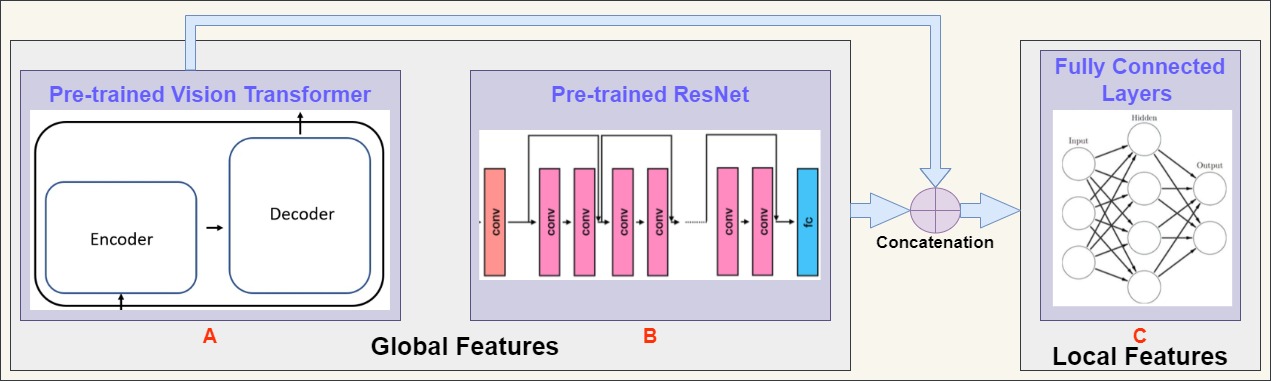}
	\end{center}
	\caption{Block diagram of backbone architecture with ViT, ResNet, and FC layers.}
	\label{ViT}
\end{figure}

The proposed backbone architecture combines a pre-trained ViT model and ResNet-18 to construct our Prototypical Meta-Learning Networks. The pre-trained ViT model extracts visual features, leveraging its transformer-based representation to capture global context. Simultaneously, the ResNet-18 processes images with pre-trained frozen weights to capture downstream task-specific additional features. By freezing the weights of both models during training, the architecture maintains the generalization abilities of the pre-trained models while focusing fine-tuning efforts on the two FC layers for downstream tasks. We concatenate the combined features from both models to create a robust representation of the input, then refine it through FC layers with batch normalization and ReLU activation. This hybrid approach ensures better task adaptation by leveraging convolutional (localized) and transformer-based (global) features, resulting in a more robust and generalized FSL prototype meta-learning network. Fig. \ref{ViT} shows the block diagram of the backbone architecture. Parts A and B represent the pre-trained ViT and ResNet-18 models, respectively. Part C of this figure indicates the FC layers, which we trained (fine-tuned) using our customized dataset to extract local features. The global features from ViT and ResNet are first concatenated and then fed to the FC layers.
\subsection{Proposed Few-shot Prototype Meta-learning Networks}
The few-shot prototypical meta-learning network maps each class to a prototype, the mean of the feature vectors of the support set examples, and classifies by measuring the distance between new queries and these prototypes \cite{snell2017prototypical}. In our architecture, support set images pass through ResNet-18 and ViT as we extract task-specific and global features, which we combine into class prototypes. During inference, query features are extracted similarly and compared to class prototypes using a pairwise distance metric. Here, distance is measured using Euclidean distance, which measures the straight-line distance between each query feature vector and each prototype feature vector, as in (\ref{Eqn:1}).
\begin{equation}
\label{Eqn:1}
        d(q, p) = \sqrt{\sum_{i=1}^{n} (q_i - p_i)^2}
\end{equation}

\begin{figure}
	\begin{center}
		\includegraphics[width=\linewidth]{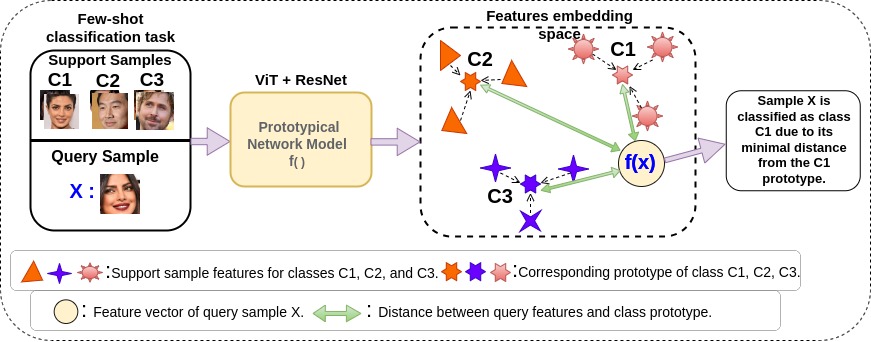}
	\end{center}
	\caption{The architecture of the designed prototype network first generates support sample features (C1, C2, C3). These support features and query samples are then passed through the ViT+ResNet backbone to generate prototype features. The query's class is subsequently determined based on the Euclidean distance to the prototype.}
\label{Prototype}
\end{figure}

Here $q$ represents the combined query features and $p$ represents the combined class prototypes.
The model predicts the class label based on the closest prototype in the feature space, enabling it to generalize to new classes with limited examples, making it ideal for few-shot learning tasks. This approach is simple and effective, making it well-suited for scenarios with limited data such as biometric face authentication \cite{yang2023two}. The architecture of our network is depicted in Fig. \ref{Prototype}.

\section{Experiment}
\label{Exp}
\subsection{Dataset}
\textbf{Customized Dataset for Fine-tuning}: The experimental dataset for the proposed model is created by collecting data from two reliable sources: ``Labelled Faces in the Wild (LFW) Dataset\footnote{https://www.kaggle.com/datasets/jessicali9530/lfw-dataset}'', and ``Pins Face Recognition\footnote{https://www.kaggle.com/datasets/hereisburak/pins-face-recognition}''.  \\  \\
\textbf{Our Generated Demographic Support and Query Face Image dataset}:
We have created support and query face image datasets considering demographic diversity for this research. We used a web scraping API to retrieve and process image data from publicly accessible sources. 
The process includes:\\ \\
\small{\textbf{API Setup --> Search Query with Key Values --> Image Retrieval --> Directory Creation --> Image Download --> Manual Filtration --> Categorization --> Pre-processing with OpenCV --> Feeding to the Networks for Testing.}} \\ \\
The dataset includes three demographic categories: Race/Ethnicity, Gender, and Age Group. The Race/Ethnicity category consists of representative faces from seven distinct racial/ethnic groups. The Gender category consists of representative faces from male, female, and non-binary. The Age Group category consists of representative faces from three groups: children, young adults, and older adults. Initially, we collected 100 images for each of the 13 classes in all three categories and fliers to around 20 to 50 face images for each of the classes. We aimed to maximize demographic diversity across all of our 
 three observed categories. We may allow the use of the dataset for non-commercial research and education purposes.  
\subsection{Pre-processing of Data}
In our pre-processing pipeline, we resize the images to $224\times224$ pixels. We then applied augmentations, including rotations of up to 30 degrees, horizontal and vertical flips with a 50\% probability each, and adjustments to brightness, saturation, contrast, and hue to handle lighting and color variations. Additionally, we performed slight translations with random shifts of up to 10\%. Finally, the images were converted to a tensor and normalized using the ImageNet dataset's mean and standard deviation values. Initially, we carefully curated the dataset to enhance the training process by limiting the selection to 40 images per class while excluding classes with less than 20 images.  
\subsection{Training}
We used a few-shot meta-learning approach for face authentication with only a few image samples. The experiment had three phases, each with a different ViT model, but all shared the same hyperparameter values. We froze the ViT and ResNet module's pre-trained weights, focusing on fine-tuning the fully connected (FC) block. This process improved task-specific performance.

Models were trained for 250 episodes using a customized dataset with a $k$-way, $n$-shot protocol, where $k$ and $n$ are randomly chosen between 5 to 8 and 1 to 5, respectively. Each episode involves 25 mini-epochs. We fine-tuned only the FC layers with 20\% dropout and L2 regularization (1e-4). The Cosine annealing learning rate scheduler adjusts the learning rate, while gradient norms were clipped at 1.0 to prevent exploding gradients. The loss was computed using cross-entropy loss and updated with the Adam optimizer.

The model learned an adaptable representation for new tasks using a dataset of 1036 classes. We labeled image pairs with dissimilarity scores (0 for the same class, 1 for different classes). The training used binary cross-entropy loss as in (\ref{eqn:2}).
\begin{equation}
 L(\hat{y}, y)= -(ylog(\hat{y})+(1 - y)\mbox{log}(1 - \hat{y}))
 \label{eqn:2}
\end{equation}
Here $\hat{y}$ is the predicted similarity/dissimilarity score and $y$ is the true label.

Training and validation losses decreased steadily but with some fluctuation in the learning rate: le-4. We used early stopping with a separate validation set of 1036 classes. Validation loss drops quickly at first and then slows down. Some overfitting was noticed around 300 epochs, so we stopped early with a training loss of about 0.56 and a validation loss of about 0.65. 
\begin{table}[!b]\centering
\caption{Performance results of our prototype networks with different ViT backbones and ResNet-18 on our demographic dataset. The data represent the mean values of 10 independent trials along with the standard deviation, denoted by the plus-minus symbol.}\label{Table1}
\resizebox{\textwidth}{!}{
\scriptsize
\begin{tabular}{|c|c|cccc|c|}\hline
\textbf{Shot} &\textbf{ViT Model}& \textbf{Metrics} &\textbf{Race} &\textbf{Gender} &\textbf{Age} \\\hline
\multirow{9}{*}{1-Shot} &\multirow{3}{*}{Facebook's DeiT} &\textbf{Accuracy} & 52.29 ± 0.03 & 83.33 ± 0.03 & 86.67 ± 0.02 \\ 
& &\textbf{Precision} & 58.58 ± 12.69 & 85.28 ± 4.22 & 88.77 ± 1.66 \\ 
& &\textbf{Recall} & 52.29 ± 3.59 & 83.33 ± 3.33 & 86.67 ± 2.36 \\\cline{2-6}
&\multirow{3}{*}{Google's VT} &\textbf{Accuracy} & 58.00 ± 1.63 & 92.00 ± 1.83 & 90.67 ± 1.49 \\ 
& &\textbf{Precision} & 62.45 ± 3.34 & 92.95 ± 1.46 & 91.83 ± 1.56 \\ 
& &\textbf{Recall} & 58.00 ± 1.63 & 92.00 ± 1.83 & 90.67 ± 1.49 \\\cline{2-6}
&\multirow{3}{*}{Microsoft's ST} &\textbf{Accuracy} & 64.00 ± 2.12 & 91.33 ± 1.83 & 94.67 ± 1.83 \\ 
& &\textbf{Precision} & 68.42 ± 4.72 & 91.99 ± 2.24 & 95.01 ± 1.85 \\ 
& &\textbf{Recall} & 64.00 ± 2.12 & 91.33 ± 1.83 & 94.67 ± 1.83 \\\cline{1-6}
\multirow{9}{*}{3-Shot} &\multirow{3}{*}{Facebook's DeiT} &\textbf{Accuracy} & 58.29 ± 2.56 & 86.00 ± 1.49 & 86.00 ± 2.79 \\ 
& &\textbf{Precision} & 66.62 ± 5.78 & 88.34 ± 2.04 & 87.71 ± 2.17 \\ 
& &\textbf{Recall} & 58.29 ± 2.56 & 86.00 ± 1.49 & 86.00 ± 2.79 \\\cline{2-6}
&\multirow{3}{*}{Google's VT} &\textbf{Accuracy} & 69.71 ± 2.93 & 93.33 ± 4.71 & 95.33 ± 1.83 \\ 
& &\textbf{Precision} & 72.46 ± 4.12 & 93.91 ± 4.34 & 95.74 ± 1.73 \\ 
& &\textbf{Recall} & 69.71 ± 2.93 & 93.33 ± 4.71 & 95.33 ± 1.83 \\\cline{2-6}
&\multirow{3}{*}{Microsoft's ST} &\textbf{Accuracy} & 80.57 ± 2.17 & 97.33 ± 1.49 & 97.33 ± 1.49 \\ 
& &\textbf{Precision} & 82.26 ± 3.26 & 97.58 ± 1.36 & 97.58 ± 1.36 \\ 
& &\textbf{Recall} & 80.57 ± 2.17 & 97.33 ± 1.49 & 97.33 ± 1.49 \\\cline{1-6}
\multirow{9}{*}{5-Shot} &\multirow{3}{*}{Facebook's DeiT} &\textbf{Accuracy} & 62.29 ± 1.28 & 87.33 ± 2.79 & 90.00 ± 2.36 \\ 
& &\textbf{Precision} & 66.07 ± 4.71 & 89.14 ± 2.74 & 91.72 ± 1.70 \\ 
& &\textbf{Recall} & 62.29 ± 1.28 & 87.33 ± 2.79 & 90.00 ± 2.36 \\\cline{2-6}
&\multirow{3}{*}{Google's VT} &\textbf{Accuracy} & 72.00 ± 1.63 & 94.67 ± 1.83 & 99.33 ± 1.49 \\ 
& &\textbf{Precision} & 74.22 ± 0.80 & 95.23 ± 1.65 & 99.39 ± 1.36 \\ 
& &\textbf{Recall} & 72.00 ± 1.63 & 94.67 ± 1.83 & 99.33 ± 1.49 \\\cline{2-6}
&\multirow{3}{*}{Microsoft's ST} &\textbf{Accuracy} & 88.29 ± 2.35 & 98.00 ± 1.83 & 99.33 ± 1.49 \\ 
& &\textbf{Precision} & 89.35 ± 2.59 & 98.18 ± 1.66 & 99.39 ± 1.36 \\ 
& &\textbf{Recall} & 88.29 ± 2.35 & 98.00 ± 1.83 & 99.33 ± 1.49 \\\hline
\end{tabular}
}
\end{table}
\subsection{Testing}
As mentioned, we evaluated all the trained models on three custom demographic dataset categories into Race/Ethnicity, Gender, and Age Group. We tested those models on three different shot settings: 1-shot, 3-shot, and 5-shot. Each testing assignment has 10 static query images for each class. Our study used a systematic sampling strategy to ensure consistency in different $n$-shot scenarios. For example, with Race/Ethnicity demographic categories, there are a total of 7 distinct classes, so in a 1-shot test scenario, we choose one static support sample image for each of the classes. As the number of shots increases (e.g., 3-shot, 5-shot), we keep the initial support images and utilize additional support images. Each data point in the performance table (Table \ref{Table1}) is derived from five different test assignments, with the results presented as the mean and standard deviation to ensure accurate and structured findings.
\section{Result and Discussion}
\label{RD}
Table \ref{Table1} presents 10 independent observations of the performance results of the prototype meta-learning networks using different ViT backbones (Facebook DeiT, Google VT, and Microsoft ST) along with ResNet-18 on our demographic dataset across three demographic factors: race/ethnicity, gender, and age. We recorded accuracy, precision, and recall for each demographic category under 1-shot, 3-shot, and 5-shot scenarios.\\ \\
\textbf{1-Shot Learning}: Microsoft's ST consistently outperforms Facebook's DeiT and Google's VT, with the highest accuracy and precision of 64.00\% for race, 91.33\% for gender, and 94.67\% for age. For the Precision metric, Microsoft's ST achieved the highest among the others for Race and Age with 68.42\% and 95.01\%, respectively, but Google's VT achieved the best for the gender group with 92.95\%. \\ \\
\textbf{3-Shot Learning}: In this scenario, Microsoft's ST again maintains superior performance, achieving 80.57\% accuracy for race and 97.33\% for both gender and age. In this scenario, Google's VT also shows good results, particularly in gender and age, with 93.33\% and 95.33\% accuracy, respectively. For the Precision metric, Microsoft's ST achieved the highest in all three categories with 82.26\%, 97.58\%, and 97.58\% respectively.\\ \\
\textbf{5-Shot Learning}: In this scenario, also Microsoft's ST exhibits the best overall performance, reaching 88.29\% accuracy for race and near-perfect scores for gender and Age (98.00\% and 99.33\%, respectively). Google's VT also performs well, especially for Age, where it attains 99.33\% accuracy. For the precision metric, Microsoft's ST achieved the highest in all three categories with 89.35\%, 98.18\%, and 99.39\% respectively, and Google's VT achieved 99.39\% for Age. 

 We observed similar patterns for recall as accuracy in all three shot scenarios. Overall, Microsoft's ST backbone provides the most stable and accurate results across the demographic categories and shot counts, especially in gender and age classification.  
\begin{figure} [!t]
	\begin{center}		\includegraphics[width=\linewidth]{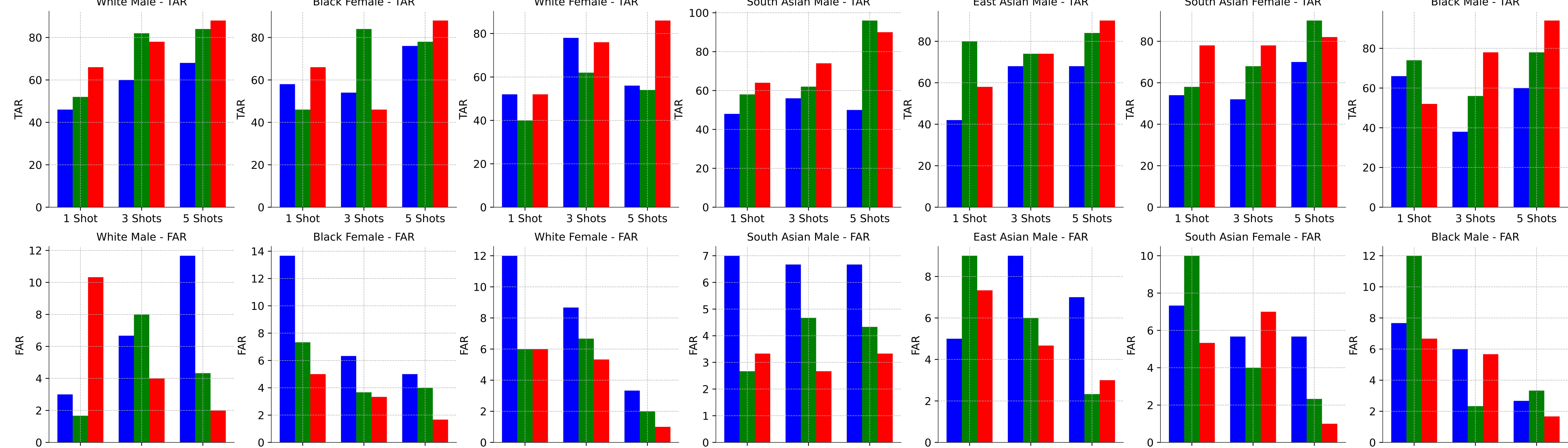}
	\end{center}
	\caption{True Authentication Rate (TAR) and False Authentication Rate (FAR) on Race/Ethnicity Category for 1-shot, 3-shot and 5-shot.}
	\label{RaceGraph}
\end{figure}
\begin{figure} [!b]
	\begin{center}
		\includegraphics[width=\linewidth]{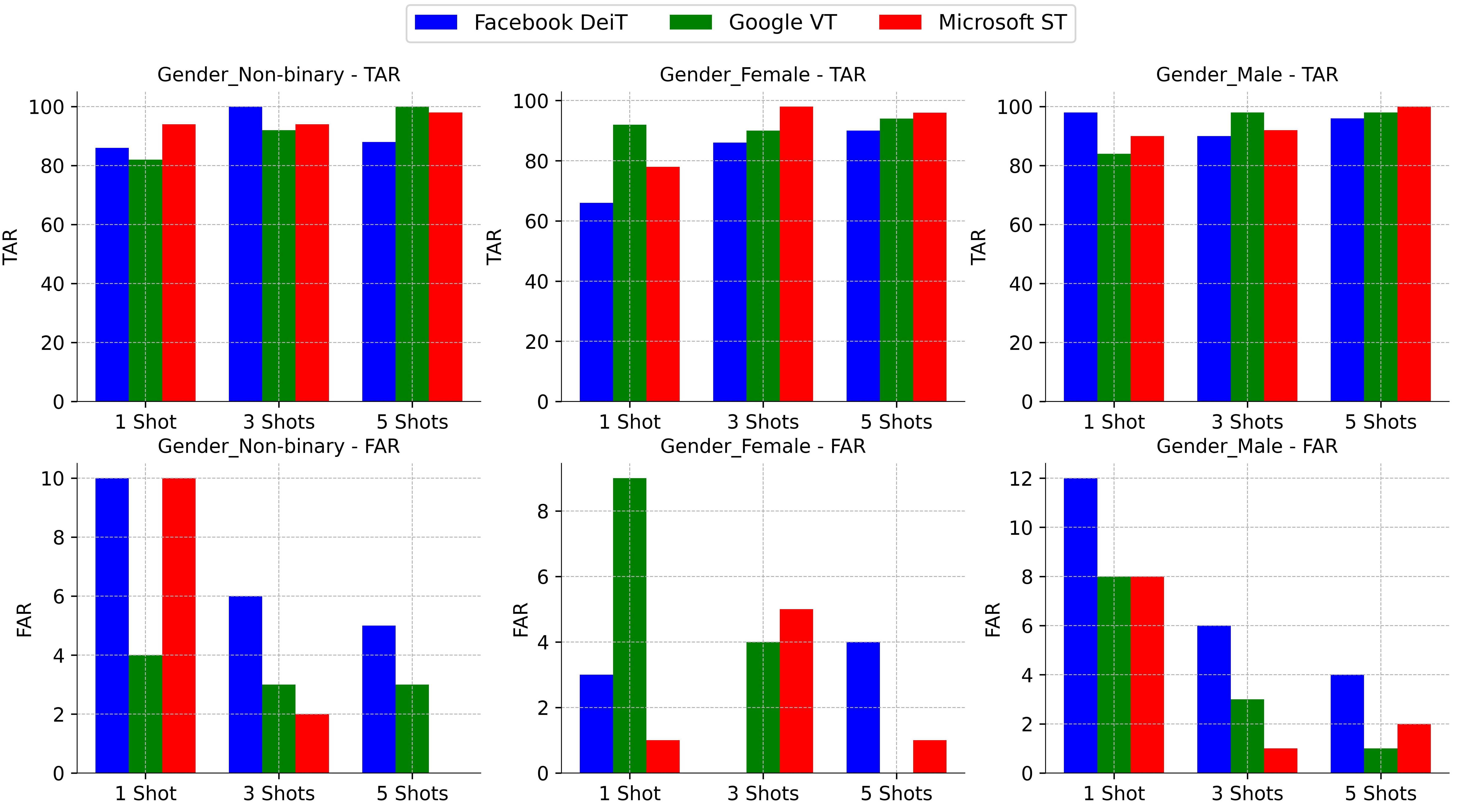}
	\end{center}
	\caption{True Authentication Rate (TAR) and False Authentication Rate (FAR) on Age Group Category for 1-shot, 3-shot and 5-shot.}
	\label{GenderGraph}
\end{figure}
\begin{figure}
	\begin{center}
		\includegraphics[width=\linewidth]{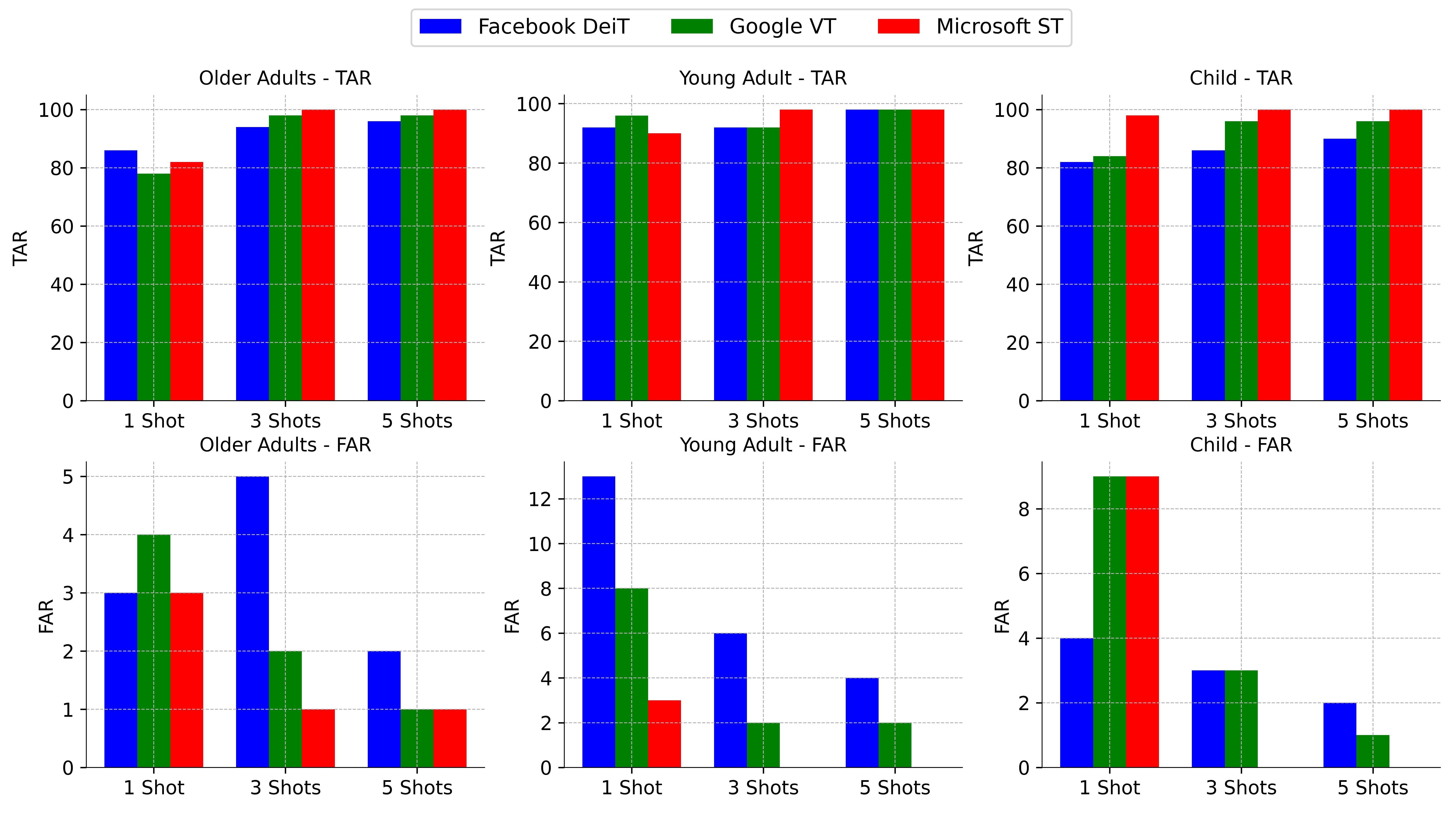}
	\end{center}
	\caption{True Authentication Rate (TAR) and False Authentication Rate (FAR) on Gender Group Category for 1-shot, 3-shot and 5-shot.}
	\label{AgeGraph}
\end{figure}
\begin{equation}
  TAR = TrueNegative / (FalsePostitive + TrueNegative)
  \label{eqn:3}
\end{equation}
\begin{equation}
FAR = FalsePostitive / (FalsePostitive + TrueNegative) 
  \label{eqn:4}
\end{equation}
The class-wise mean True Authentication Rate (TAR) and False Authentication Rate (FAR) 
which are calculated in (\ref{eqn:3}) and in (\ref{eqn:4}) respectively, are displayed in graphs in Fig. \ref{RaceGraph}, Fig. \ref{GenderGraph}, and Fig. \ref{AgeGraph}, show clear trends that as the support set increases, TAR improves, and FAR decreases across all backbone networks.
Microsoft ST (in red) demonstrates the most stable performance on these TAR and FAR metrics. Some biases remain when testing a demographically diverse face dataset, whereas few are likely due to our data representation issues. However, Microsoft ST shows the most negligible bias, followed closely by Google VT. If we explore the graphs in Fig. \ref{GenderGraph}, we see that all three models performed well and showed minimal bias in the TAR metrics for this demographic class. However, regarding the FAR metric, Microsoft ST is again the clear winner, followed by Google ViT. From the graphs in Fig. \ref{AgeGraph}, the same conclusion could be drawn on performances on the age group dataset.  

Considering all the metrics used—accuracy, precision, recall, class-wise TAR, and FAR—our empirical study shows that Microsoft ST outperforms the other two models, followed by Google ViT. 

\begin{figure}[!b]
	\begin{center}
		\includegraphics[width=0.99\linewidth]{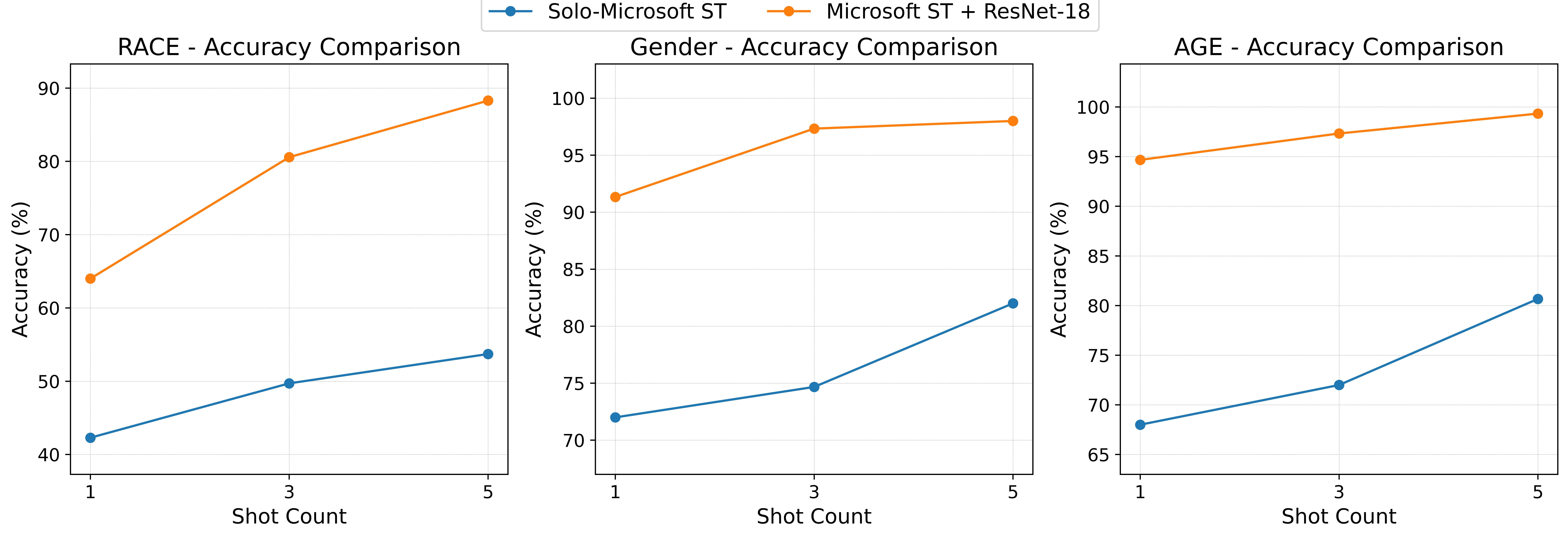}
	\end{center}
	\caption{Ablation study: Mean accuracy of Microsft ST alone in blue lines and ViT+ ResNet-18 in orange lines in three different shots in all three demographic categories.}
	\label{AblataionGraph}
\end{figure}
\section{Ablation Study}
\label{Ablation}
We conducted an ablation study to understand the performances of only the ViT backbone without the support of ResNet. We have found that the Microsoft ST performance is better than the other two ViT backbones, so we observed an ablation study of this ViT (Due to page length constraints, we limited this ablation study to only Microsoft ST. However, an ablation study of all three ViT models could be conducted in future work to gain deeper insights into the models). We trained (fine-tuned) the FC layers and recorded the results of the Microsoft ST backbone alone in our few-shot prototype networks. We plotted the comparative mean accuracies across all three categories in the graphs shown in Fig. \ref{AblataionGraph}. Here, we observe that ViT alone did not perform well compared to when we combined it with ResNet-18.
\section{Conclusion}
\label{Conclusion}
In this paper, we conducted an empirical study to explore the efficacy of Vision Transformers (ViT) supported by ResNet in demographic face authentication tasks. The focus was on leveraging global features while minimizing reliance on local features. By integrating pre-trained ViT's and ResNet's global features, we fine-tuned the last two fully connected layers for local features. We designed a novel few-shot prototype meta-learning network to conduct this experiment on our created demographic face authentication task. Our study used three state-of-the-art ViT models (Facebook DeiT, Google VT, and Microsoft ST) and demonstrated that combining global and local features consistently improves performance across different demographic groups, including race, gender, and age, from one-shot to five-shot.

The results also indicate that ViT's global features alone can help authenticate faces. However, adding features from the ResNet layer significantly enhances accuracy and generalization, especially in limited training sets. Notably, the model maintained consistent performance across diverse demographic groups, reducing potential biases in face authentication tasks. Among the models evaluated, Microsoft's ST exhibited better performance, particularly in few-shot learning conditions, suggesting that this combination of architectures is more robust in few-shot scenarios.

Overall, this empirical study contributes to the ongoing efforts to improve fairness in demographic face authentication systems using advanced deep learning architectures like ViT. Future work will focus on refining the network's ability to generalize across even more diverse datasets and exploring additional architectures that can improve computational efficiency and accuracy without sacrificing fairness.

\section*{Acknowledgment}
This research was partially supported by the project Future Artificial Intelligence Research FAIR CUP B53C220036 30006 grant number PE0000013. \\ \\
The authors thank Arturo Argentieri from CNR-ISASI, Italy, for his technical contributions to the multi-GPU computing facilities.

%
%
\bibliographystyle{splncs04}

\bibliography{Ourbib}

\end{document}